\documentclass[sn-standardnature]{sn-jnl}

\jyear{2023}%

\unnumbered

\usepackage{caption}
\usepackage{subcaption}
\usepackage{float} 
\usepackage{pifont}
\newcommand{\cmark}{\ding{51}}%
\newcommand{\xmark}{\ding{55}}%

\usepackage{wrapfig}

\begin{document}

\title[Redefining Digital Health Interfaces with LLMs]{Redefining Digital Health Interfaces with Large Language Models}

\author*[1]{\fnm{Fergus} \sur{Imrie}}\email{imrie@ucla.edu}
\author[2]{\fnm{Paulius} \sur{Rauba}}\email{pr501@cam.ac.uk}
\author[2,3]{\fnm{Mihaela} \sur{van der Schaar}}\email{mv472@cam.ac.uk}

\affil[1]{\orgname{University of California, Los Angeles}, \orgaddress{\country{USA}}}
\affil[2]{\orgname{University of Cambridge}, \orgaddress{\country{UK}}}
\affil[3]{\orgname{The Alan Turing Institute}, \orgaddress{\country{UK}}}

\abstract{
Digital health tools have the potential to significantly improve the delivery of healthcare services. However, their adoption remains comparatively limited due, in part, to challenges surrounding usability and trust.
Large Language Models (LLMs) have emerged as general-purpose models with the ability to process complex information and produce human-quality text, presenting a wealth of potential applications in healthcare. 
Directly applying LLMs in clinical settings is not straightforward, however, with LLMs susceptible to providing inconsistent or nonsensical answers. 
We demonstrate how LLM-based systems can utilize external tools and provide a novel interface between clinicians and digital technologies. 
This enhances the utility and practical impact of digital healthcare tools and AI models while addressing current issues with using LLMs in clinical settings such as hallucinations.
We illustrate LLM-based interfaces with the example of cardiovascular disease risk prediction. We develop a new prognostic tool using automated machine learning and demonstrate how LLMs can provide a unique interface to both our model and existing risk scores, highlighting the benefit compared to traditional interfaces for digital tools.
}


\maketitle

\section{Introduction}\label{sec:intro}

Digital healthcare technologies represent a frontier in medicine. Despite a multitude of tools being developed \cite{sutton2020overview,dunn2018wearables}, clinical adoption of such methods faces significant hurdles \cite{eichler2007barriers,Mathews2019}, with some even calling their use ``infeasible'' \cite{Muller2010} and ``substantially conceptual'' \cite{abernethy2022promise}.
One key issue is usability \cite{Ratwani2019decade}, which can result in errors associated with patient harm \cite{Howe2018electronic} and contribute to clinician frustration, jeopardizing patient safety \cite{Shanafelt2016,gardner2019physician}.
New tools employing artificial intelligence (AI) and machine learning offer substantial promise, with their impact expected to be felt across all areas of healthcare \cite{Bajwae2021}. Yet these approaches face the same usability challenges as existing digital tools, while presenting additional questions around model trust \cite{Rajpurkar2022,asan2020artificial}. Consequently, these issues have limited the uptake and impact of AI technologies in clinical settings thus far \cite{goldfarb2022ai,Davenport2019,Kelly2019}.

To improve the usability of clinical predictive models, several approaches have sought to simplify or automate the process of obtaining a prediction. These include points-based scoring systems \cite{Gage2001}, web-based calculators \cite{Hippisley-Cox2017CVD,Imrie2023AutoPrognosis}, or integration within electronic health records \cite{Rothman2013}.
While this can make such tools easier to use, simply obtaining a prediction is frequently insufficient and more is required to build model trust with both clinicians \cite{Rajpurkar2022} and regulators \cite{FDA2019,Mourby2021}.

Large Language Models (LLMs) offer a potential solution to the challenges faced by digital tools.
LLMs have recently captured the imagination of both the research community and the general public, pushing the boundaries of human-machine interaction.
Consequently, there is great interest in applying LLMs in healthcare, with potential applications including facilitating clinical documentation, summarizing research papers, or as a chatbot for patients \cite{Moor2023}.

Applying LLMs in safety-critical clinical settings is not straightforward.
LLMs may provide inconsistent or nonsensical answers \cite{Singhal2023,Lecler2023revolutionizing} and have a tendency to hallucinate facts \cite{maynez2020faithfulness,Ji2023survey}.
This is clearly unacceptable in medicine when making high-stakes decisions. 
Additionally, LLMs can encounter difficulty with seemingly basic functionality, such as mathematical calculations or factual lookup \cite{patel2021nlp,schick2023toolformer}, and are unable to access up-to-date information by default \cite{komeili2022internet}.
These limitations constrain the clinical utility of LLMs.

\begin{figure*}[t!]
  \centering
  \includegraphics[width=\textwidth,trim=11.5em 19.5em 7.5em 13em, clip]{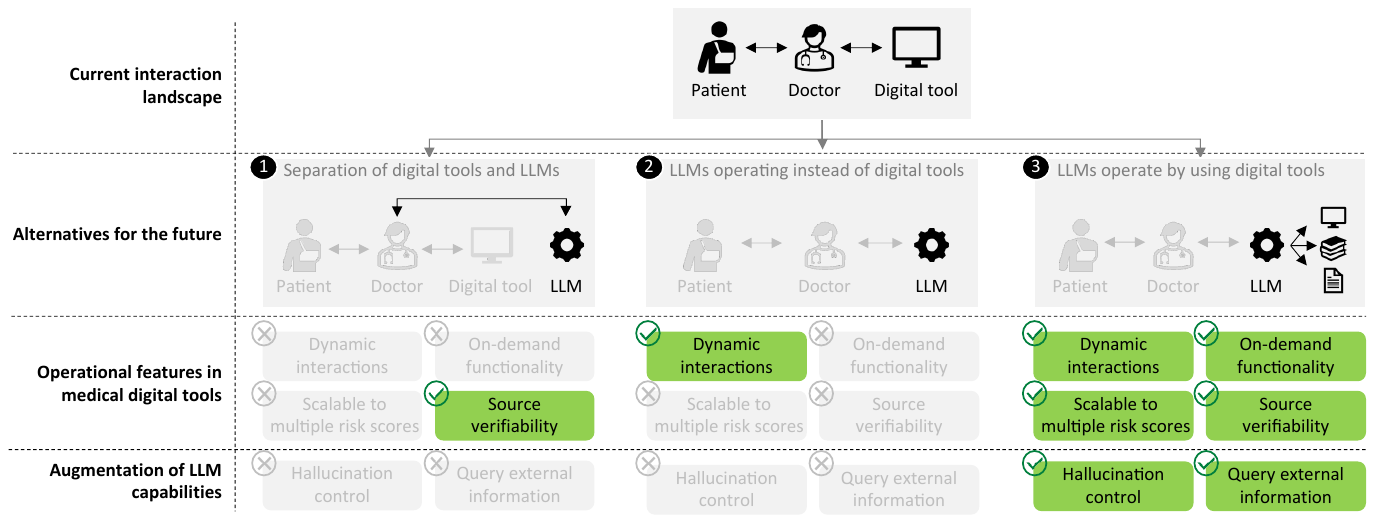}
  \caption{Clinicians have previously needed to interact directly with digital tools, such as risk scores. While others have discussed LLMs replacing existing clinical tools (1,2), we envisage LLMs forming a novel interface by enabling dynamic interactions and facilitating deeper engagement with tools and related information, such as model explainability, medical papers, and clinical guidelines (3).}
  \label{fig:interface}
  \vspace{0.1cm}
\end{figure*}

In this paper, we explore a new application of LLMs in healthcare and propose using LLMs to facilitate clinician interactions with AI models and digital tools. Conceptually, this differs substantially from previous applications, such as training medicine-specific LLMs \cite{Luo2022,Yang2022large} or using LLMs for prediction \cite{Jiang2023health}.
Instead, we construct LLM-based systems, offering an intuitive natural language interface that can streamline clinician interactions with multiple digital tools and sources of information, improving efficiency and usability (Fig. \ref{fig:interface}).

Dynamic interactions in the form of natural language dialogues have been identified as a key feature for practitioners to deploy machine learning models in healthcare \cite{lakkaraju2022rethinking}. The most appropriate interface or dialogue cannot be pre-specified; instead, it depends on the clinician and the patient. Thus, the ability to adapt and tailor interactions represents a critical advance in the functionality of such tools that can be unlocked by LLMs.

By default, LLMs do not possess the ability to access external tools or information.
We augment the base functionality of LLMs and enable them to access approved medical tools and other sources of information, thereby not solely relying on the inherent capabilities of a given LLM.
This approach is scalable to multiple predictive models, unifying digital tools within a single, natural language-based interface. 
By adopting a systems approach, the LLM does not itself issue predictions and can access relevant domain-specific information, rather than needing to supply specific knowledge itself. 
Consequently, the potential for hallucinations is limited and we ensure actionable information is provided by approved clinical sources.

As an example of our approach, we specifically focused on risk scoring \cite{moons2009prognosis} and considered cardiovascular disease (CVD), the most common cause of mortality globally \cite{Muthiah2022}.
Primary prevention uses prognostic models to estimate the future risk of developing CVD. This allows high-risk individuals to be identified and their risk to be managed via interventions, such as lifestyle modifications or pharmaceuticals.
Numerous CVD risk scores have previously been developed, for example the Framingham score \cite{DAgostino2008} in the United States, SCORE2 \cite{SCORE2} in Europe, and QRisk3 \cite{Hippisley-Cox2017CVD} in the UK.
We begin by developing an improved risk score for CVD using AutoPrognosis 2.0 \cite{Imrie2023AutoPrognosis}, an open-source automated machine learning framework for developing diagnostic and prognostic models.
We then demonstrate a novel LLM-based interface for both our prognostic model and existing risk scores. We provide several illustrative examples of dynamic interactions that substantially extend the capabilities of existing fixed interfaces.

\begin{table*}[!ht]
\caption{\textbf{Model validation on UK Biobank.} C-indices, Brier scores, and Expected/Observed ratios for our approach (AP2), QRisk3, SCORE2, and Framingham score. We report results for the original existing risk scores and the performance following recalibration. Mean performance (95\% CI).}
\label{tbl:overall_res}
\vspace{0.1in}
\centering
\resizebox{\textwidth}{!}{
\begin{tabular}{lccc}
\toprule
                    & \textbf{C-Index} $\uparrow$ & \textbf{Brier Score} $\downarrow$ & \textbf{E/O} \\ 
\midrule
\textbf{AP2}        & 0.741 (0.740-0.743) & 0.041 (0.041-0.041) & 1.003 (1.001-1.006) \\
\midrule
\multicolumn{4}{c}{\textit{Original Risk Scores}} \\
\textbf{Framingham} & 0.705 (0.703-0.707) & 0.047 (0.047-0.047) & 2.495 (2.491-2.499) \\
\textbf{SCORE2}     & 0.710 (0.709-0.712) & 0.042 (0.042-0.042) & 1.606 (1.603-1.608) \\
\textbf{QRisk3}     & 0.722 (0.721-0.724) & 0.044 (0.043-0.044) & 1.941 (1.937-1.944) \\
\midrule
\multicolumn{4}{c}{\textit{Recalibrated Risk Scores}} \\
\textbf{Framingham} & 0.705 (0.703-0.707) & 0.041 (0.041-0.041) & 1.000 (0.998-1.002) \\
\textbf{SCORE2}     & 0.710 (0.709-0.712) & 0.041 (0.041-0.041) & 1.000 (0.998-1.002) \\
\textbf{QRisk3}     & 0.722 (0.721-0.724) & 0.041 (0.041-0.041) & 1.000 (0.998-1.002) \\
\bottomrule
\end{tabular}
}
\end{table*}

\section{Results}\label{sec:results}

We conducted experiments using data from the UK Biobank \cite{Sudlow2015}, a prospective population study of around half a million individuals from the UK, enrolled between 2006 and 2010. UK Biobank collected a broad set of information from participants at enrollment, including health and medical history, blood tests, physical examination, and socio-demographics, with ongoing linkage to healthcare records.
The outcome of interest was the incidence of CVD within a 10-year horizon, where CVD was defined as the composite endpoint of myocardial infarction (ICD-10 codes: I21, I22), angina pectoris (I20), stroke (I63, I64), or transient cerebral ischaemic attacks (G45).
The cohort contained 407,605 individuals and there were 17,600 CVD events within a 10-year horizon. A flow chart of the study cohort is provided in Fig. \ref{fig:flow_diagram} and a summary of patient characteristics is provided in Table \ref{tbl:cohort_characteristics}.
Further details can be found in Section \ref{sec:methods}.

We used AutoPrognosis 2.0 \cite{Imrie2023AutoPrognosis} to automatically construct a prognostic model for incidence of CVD.
We first validate the performance of our approach compared to existing risk scores and then demonstrate a novel interface for using such risk scores using LLMs.

\subsection{Model performance}

In Table \ref{tbl:overall_res}, we compared our prognostic model (AP2) with three existing CVD risk scores, namely QRisk3 \cite{Hippisley-Cox2017CVD}, SCORE2 \cite{SCORE2}, and Framingham score \cite{DAgostino2008}.
The ensemble model constructed using AutoPrognosis achieved a C-index of 0.741 (95\% CI: 0.740-0.743), a Brier score of 0.041 (95\% CI: 0.041-0.041), and an expected/observed (E/O) ratio of 1.003 (95\% CI: 1.001-1.006), outperforming all three existing risk scores assessed.

\begin{figure*}[ht!]
\centering
\vspace{0.5cm}
\begin{subfigure}{.49\textwidth}
  \centering
  \includegraphics[width=0.99\linewidth]{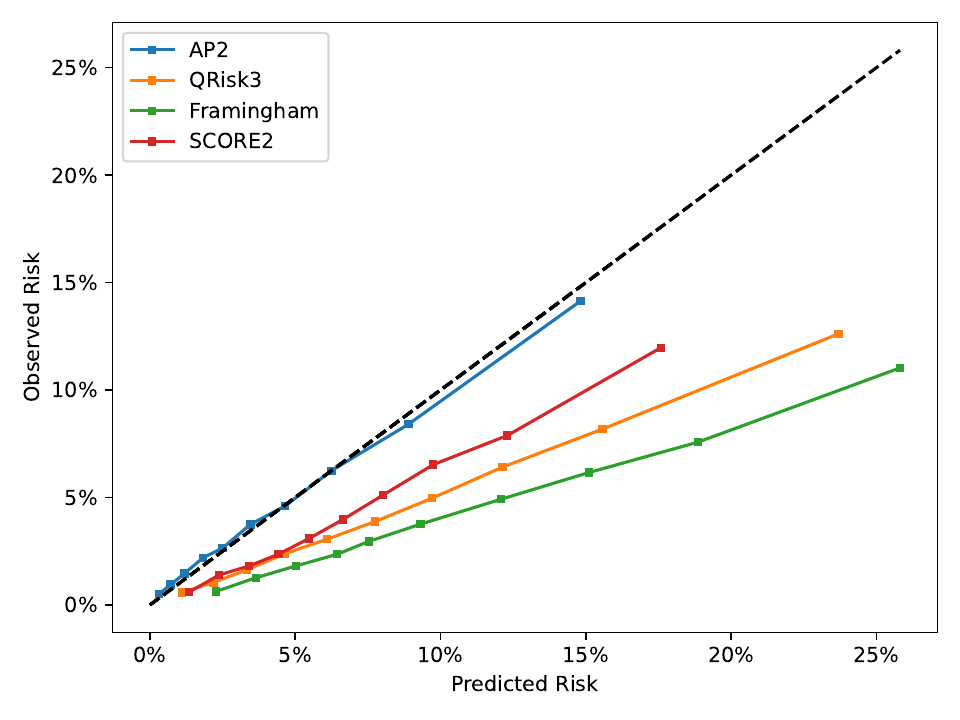}
  \caption{Original existing risk scores.}
  \label{fig:calibration_curves_uncalibrated}
\end{subfigure}\hfill
\begin{subfigure}{.49\textwidth}
  \centering
  \includegraphics[width=0.99\linewidth]{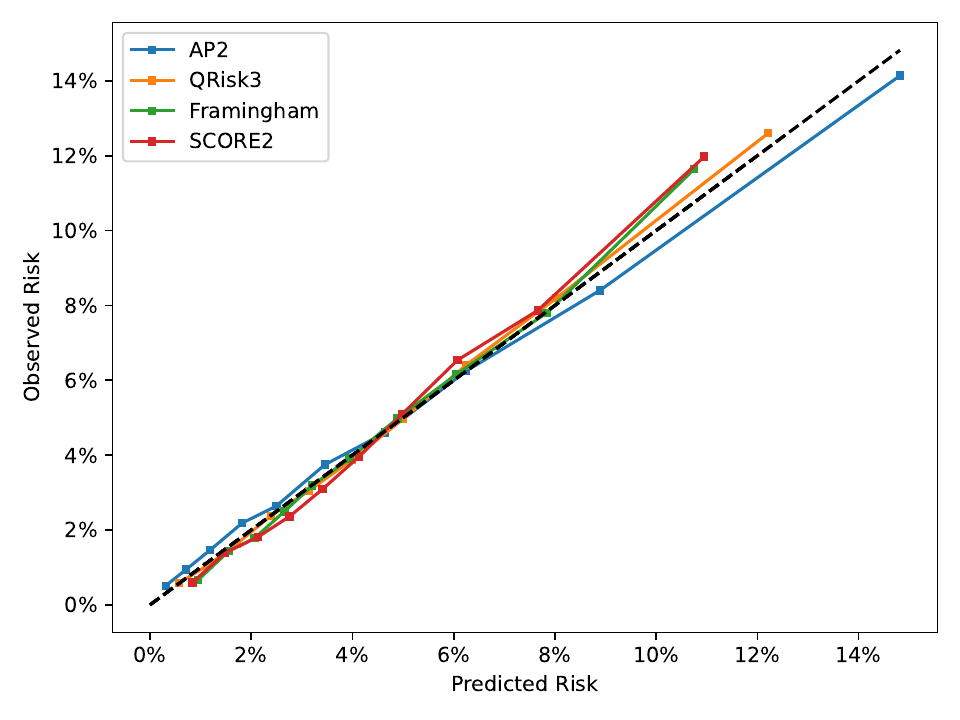}
  \caption{Recalibrated existing risk scores.}
  \label{fig:calibration_curves_recalibrated}
\end{subfigure}
\caption{\textbf{Calibration curves.} Calibration curves for our approach (AP2), QRisk3, SCORE2, and Framingham score, both before and after calibrating the existing risk scores to the UK Biobank cohort. Observed risk was calculated using Kaplan-Meier estimators \cite{kaplan1958nonparametric}.}
\label{fig:calibration_curves}
\vspace{0.25cm}
\end{figure*}

QRisk3 and SCORE2 achieved C-indexes of 0.722 (95\% CI: 0.721-0.724) and 0.710 (95\% CI: 0.709-0.712), respectively, with similar Brier scores (QRisk3: 0.044, 95\% CI: 0.043-0.044; SCORE2: 0.042, 95\% CI: 0.042-0.42). 
Both approaches overpredicted risk in the UK Biobank population (E/O ratio QRisk3: 1.941, 95\% CI: 1.937-1.944; SCORE2: 1.606, 95\% CI 1.603-1.608) consistent with previous findings \cite{Parsons2023}.
The Framingham score was the weakest performing risk score, achieving a C-index of 0.705 (95\% CI: 0.703-0.707). This underperformance likely was a consequence of the Framingham score being developed on a substantially different population from the USA.

Calibration curves (Fig. \ref{fig:calibration_curves}) show that AutoPrognosis predictions were well calibrated across all risk deciles. In contrast, the existing risk scores overpredicted risk for all individuals in the UK Biobank cohort (Fig. \ref{fig:calibration_curves_uncalibrated}). We recalibrated the existing risk scores by scaling the predictions according to the overall incidence in the UK Biobank cohort and the mean predicted risk (Fig. \ref{fig:calibration_curves_recalibrated}). Following recalibration, QRisk3 exhibited good calibration across all risk deciles, while SCORE2 and Framingham score both modestly overpredicted risk in the lower risk deciles and underpredicted risk in the highest risk decile. 

\begin{figure*}[ht!]
\centering
\vspace{0.5cm}
\begin{subfigure}{.49\textwidth}
  \centering
  \includegraphics[width=0.99\linewidth]{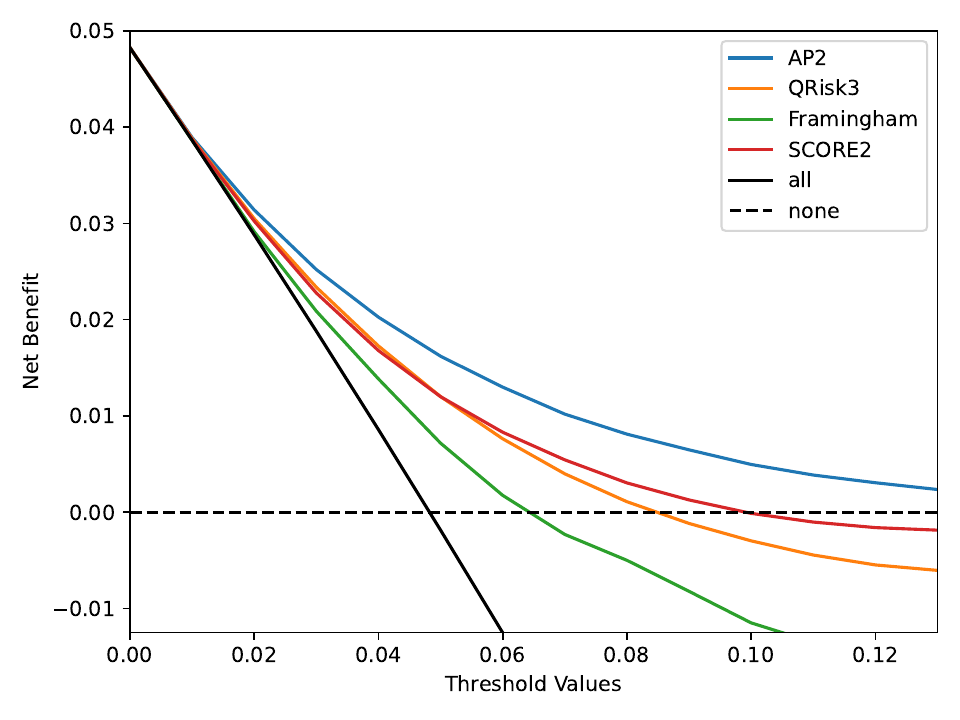}
  \caption{Original existing risk scores.}
  \label{fig:net_benefit_uncalibrated}
\end{subfigure}\hfill
\begin{subfigure}{.49\textwidth}
  \centering
  \includegraphics[width=0.99\linewidth]{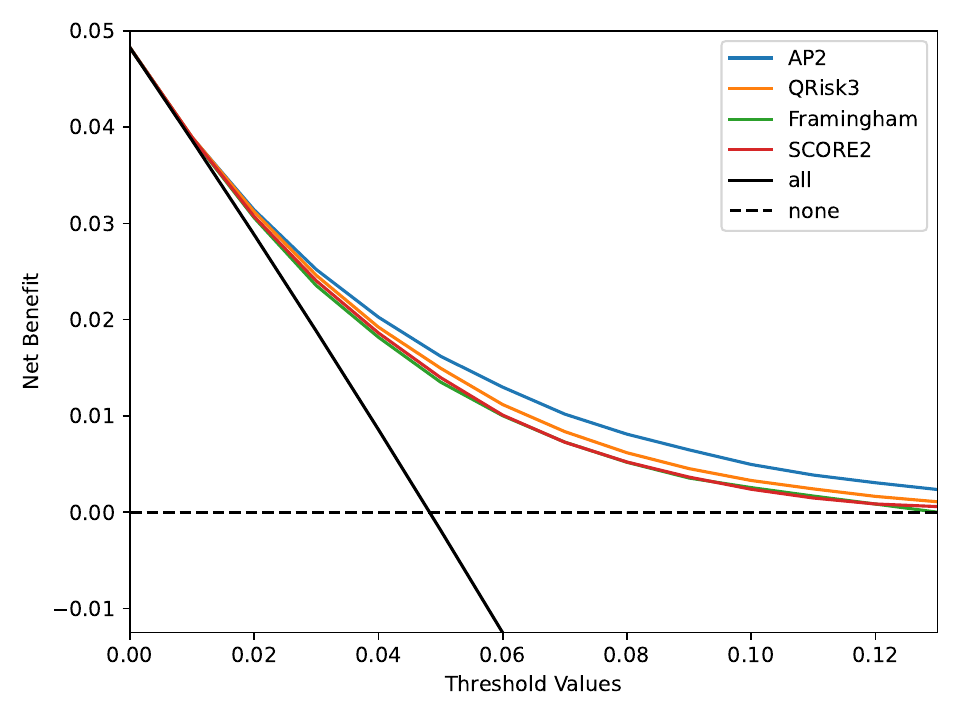}
  \caption{Recalibrated existing risk scores.}
  \label{fig:net_benefit_recalibrated}
\end{subfigure}
\caption{\textbf{Decision curve analysis.} Our approach (AP2) provides greater net benefit at all thresholds than the existing risk scores (Framingham score, SCORE2, and QRisk3) and the baseline strategies (All and None).}
\label{fig:net_benefit}
\vspace{0.25cm}
\end{figure*}

\begin{figure*}[ht!]
\centering
\includegraphics[width=\linewidth]{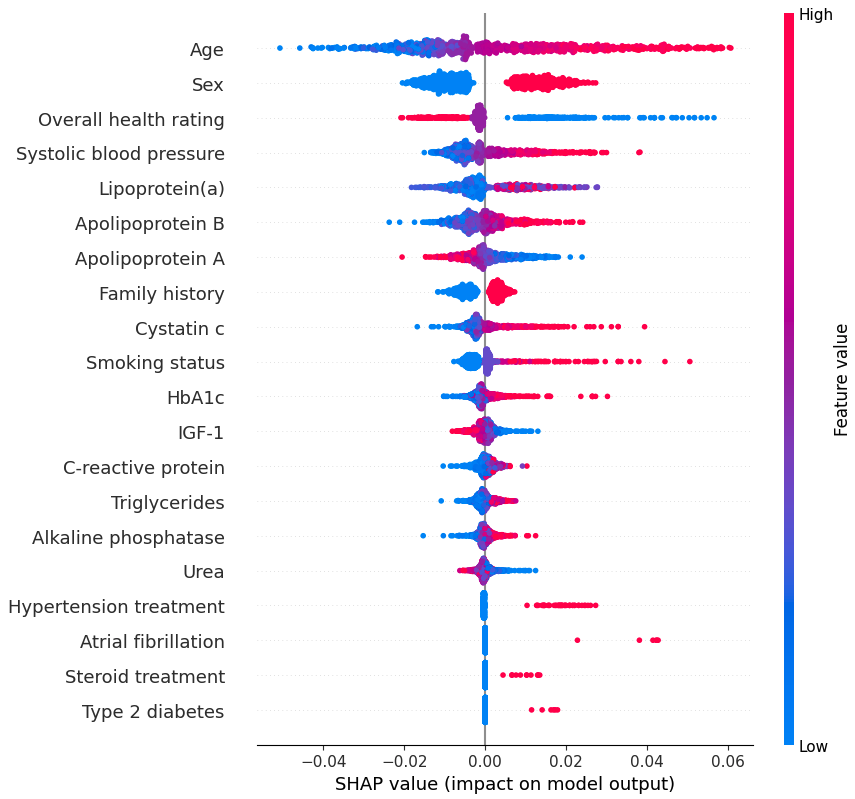}
\caption{\textbf{Feature importance.} SHAP values of the variables included in the AutoPrognosis model.}
\label{fig:feature_importance}
\vspace{-0.05cm}
\end{figure*}

To assess model performance in a manner that considers the effect on clinical decision making, we additionally performed decision curve analysis \cite{Vickers2006,Vickers2008}, as recommended in the TRIPOD guidelines \cite{Moons2015}. 
Decision curve analysis is a method to evaluate the clinical utility of a model by assessing the net benefit for a range of risk thresholds at which a decision maker would act or intervene. 
The model with the higher net benefit at any given threshold is preferred.
We compared our model with the existing risk scores, as well as baseline strategies that assume all patients will develop CVD (All) or none will (None). Our model achieved a greater net benefit at all decision thresholds compared to the existing risk scores (Fig. \ref{fig:net_benefit}).
Net benefit takes into account both discrimination and calibration. While recalibrating the existing risk scores improved their net benefit, the AutoPrognosis model still resulted in the highest net benefit at all thresholds.

To better understand the rationale for the predictions and drivers of the outperformance of our approach, we assessed feature importance by calculating SHAP values \cite{lundberg2017unified}. 
SHAP values for the 20 features included in the final version of our model are shown in Fig. \ref{fig:feature_importance}. 
Consistent with clinical knowledge, age, sex, and blood pressure are three of the most important features and were included in all three existing risk scores considered in this work. 
However, a number of additional features, including both laboratory and nonlaboratory tests, also significantly contributed to model predictions. 
A number of these features have been shown to be risk factors for CVD, but have not been incorporated into other risk scores.
For example, Apolipoprotein B (ApoB) is a primary component of several lipoproteins, including low-density lipoprotein (LDL), and transports cholesterol and other lipids in the blood.
High levels of ApoB are the primary driver of atherosclerosis and there is evidence that ApoB is a more accurate predictor of CVD than total cholesterol or LDL \cite{Marston2022,Behbodikhah2021}.
ApoB is deemed one of the most important variables in our approach (Fig. \ref{fig:feature_importance}), along with Apolipoprotein A and Lipoprotein A which have also been shown to be risk factors for CVD \cite{Erqou2010}.
Finally, ``Overall health rating'', a non-laboratory, self-reported feature, was found to be predictive, mirroring the findings of Alaa et al. regarding the benefits of non-laboratory variables for CVD risk prediction \cite{Alaa2019}.

\begin{table*}[ht]
\caption{Representative questions that a clinician might have relating to a risk score at different stages of a patient encounter, together with whether existing interfaces for risk scores provide this information. All questions can be addressed using LLM-based interfaces.}
\vspace{0.15cm}
\label{tbl:questions}
\centering
\small
\begin{tabular}{p{0.16\linewidth} p{0.63\linewidth} cp{0.04\linewidth}}
\toprule
               &                                        & \textbf{Existing} \\
\textbf{Stage} & \textbf{Representative questions}      & \textbf{interfaces} \\
\midrule
Before               & Which features does the risk score use?             & \xmark \\
Patient              & Why are these features included in the risk score?  & \xmark \\
Encounter            & How was the risk score validated?                   & \xmark \\
                     & What is the methodology underlying the risk score?  & \xmark \\
\midrule
Before Risk          & When do clinical guidelines recommend risk scoring? & \xmark \\
Scoring              & What is the recommended risk score?                 & \xmark \\
                     & Who is the risk score suitable for?                 & \xmark \\
\midrule
During Risk          & What is the risk for this patient?                  & \cmark \\
Scoring              & What characteristics led to the patient's risk?     & \xmark \\
                     & What effect would changing this feature have on the risk? & \xmark \\ 
\midrule
After Risk Scoring  & What action is recommended by the guidelines based on the risk? & \xmark \\
\bottomrule
\end{tabular}
\vspace{0.15cm}
\end{table*}

\begin{figure*}[ht!]
\centering
\includegraphics[width=\linewidth,trim=-0.5em 0em 0em 2em, clip]{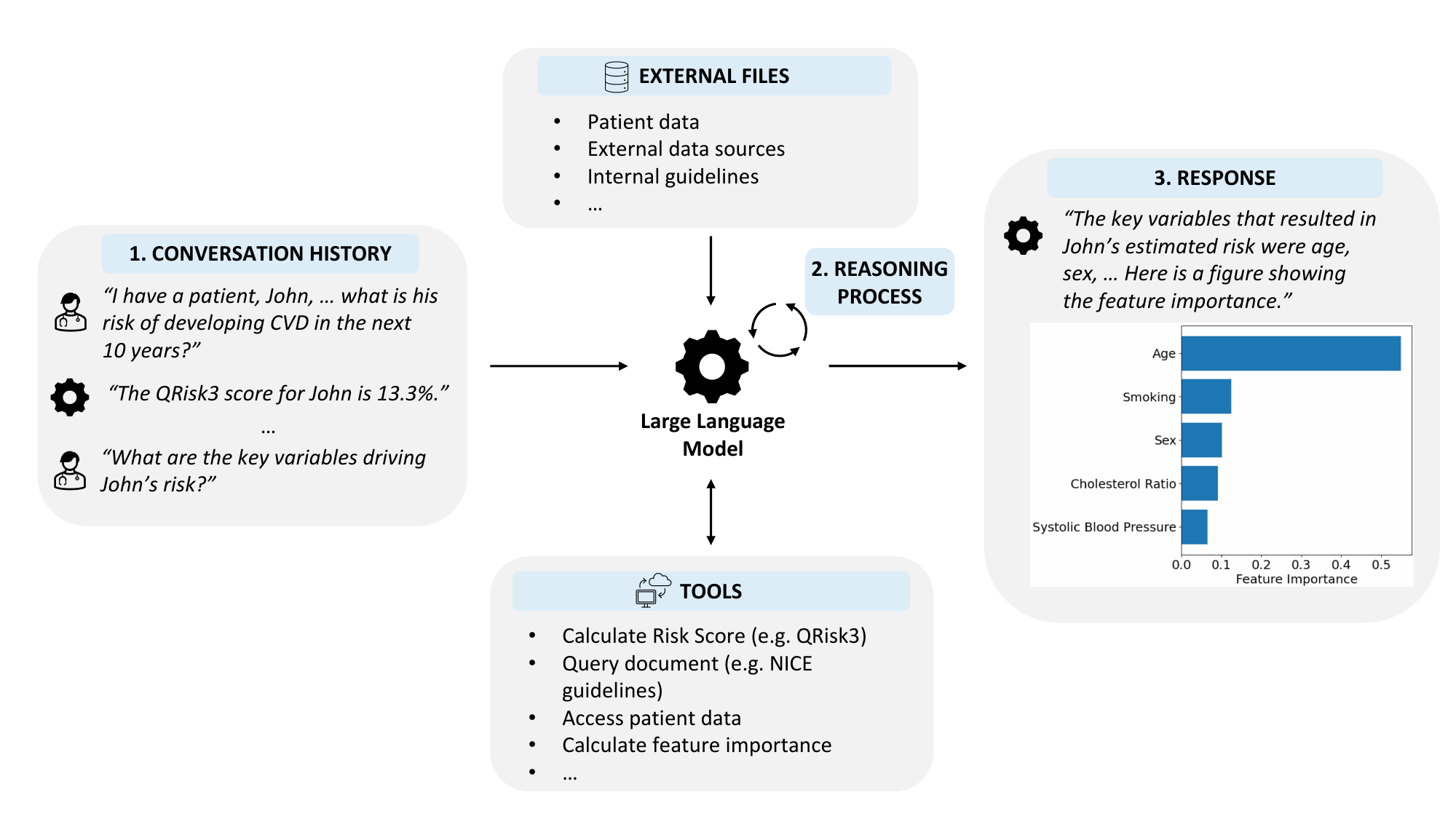}
\caption{Overview of an LLM-based system that enables clinicians to interface with digital tools using natural language inputs. (1) The LLM is provided with the history of the interaction, including the current request. (2) Using an iterative reasoning process, the LLM decides which, if any, tools are required and with what input. (3) The LLM provides a response to the user incorporating information provided by any tools that were used.}
\label{fig:systems_diagram}
\vspace{-0.05cm}
\end{figure*}

\subsection{LLM-based Interfaces}

While LLMs are powerful models for natural language processing, by default, they do not possess the ability to access external tools or information.
Methods to extend the functionality of LLMs beyond text generation are in their infancy but can already be used to significantly expand the capabilities of LLMs \cite{schick2023toolformer,nakano2021webgpt}.

In this section, we construct LLM-based systems and demonstrate multiple examples of how such systems can provide a novel interface for digital health tools, in particular clinical risk scores.
The LLM-based systems can incorporate numerous external tools, sources of information, and clinical data (illustrated in Fig. \ref{fig:systems_diagram}).
Instead of using an LLM to issue predictions directly, we augmented LLMs and enabled them to access approved medical tools and other sources of information, thereby not solely relying on the inherent capabilities of a given LLM, while unifying multiple digital tools within a single, natural language-based interface.

To demonstrate the versatility of our approach, we took an off-the-shelf pretrained LLM, GPT-4 \cite{openai2023gpt4}, and imbued it with a set of tools using in-context learning, requiring no additional training of the LLM.
Additionally, we implemented a ``source'' functionality that allows the user to see whether the LLM used a tool or accessed specific information and, if so, which tool and with what input.
This helps avoid hallucinations since it enables verification that the information was issued by an underlying clinical tool or source rather than being generated by the LLM.
Additional implementation details are provided in Section \ref{sec:methods}.

\begin{figure*}[ht!]
\centering
\vspace{0.5cm}
\begin{subfigure}{.49\textwidth}
  \centering
  \includegraphics[width=0.99\linewidth,trim=3em 32em 4em 1.30em, clip]{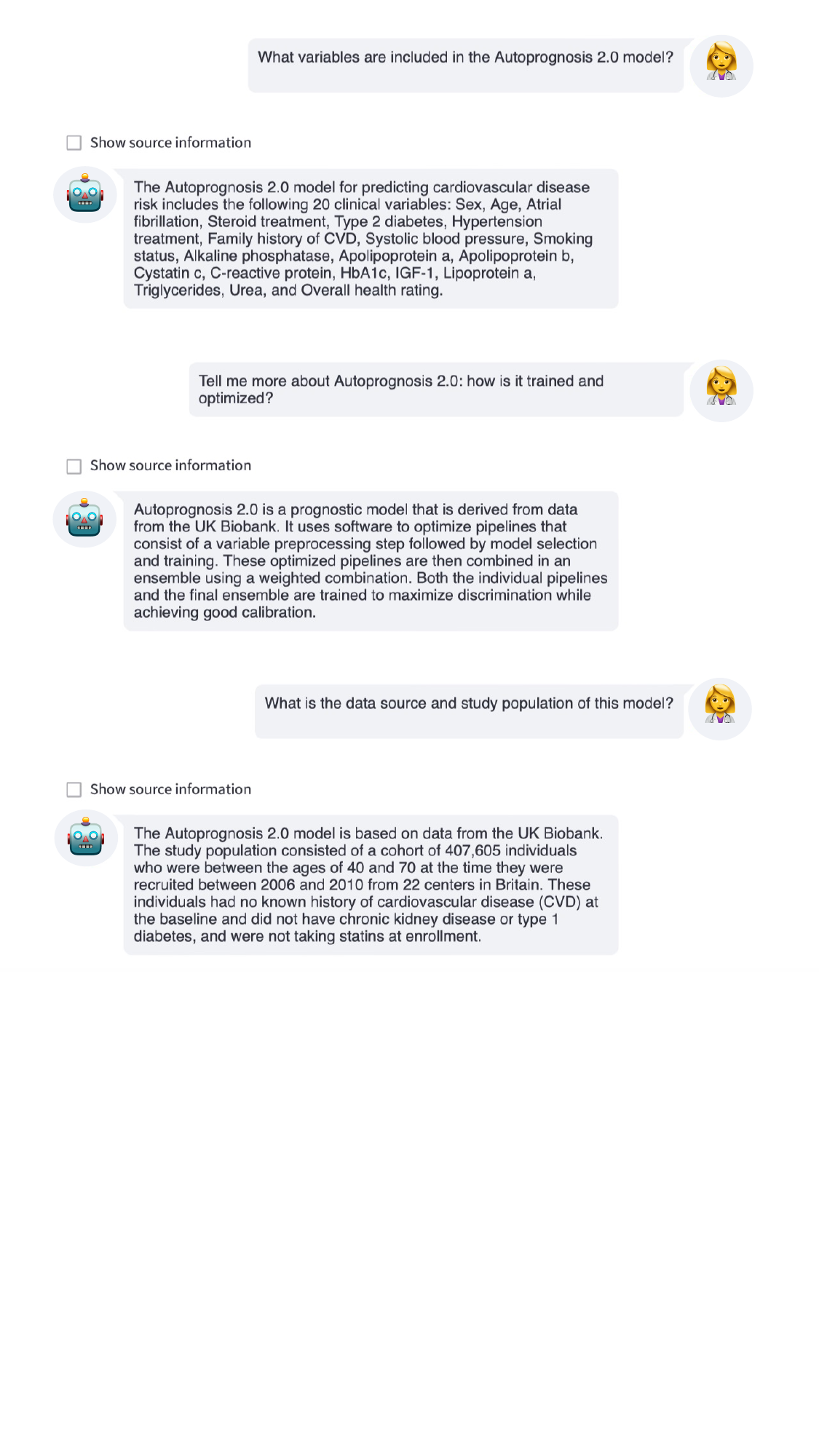}
  \caption{}
  \label{fig:eg_interaction_model_facts}
\end{subfigure}
\begin{subfigure}{.49\textwidth}
  \centering
  \includegraphics[width=0.99\linewidth,trim=2.5em 26.85em 2.3em 3.55em, clip]{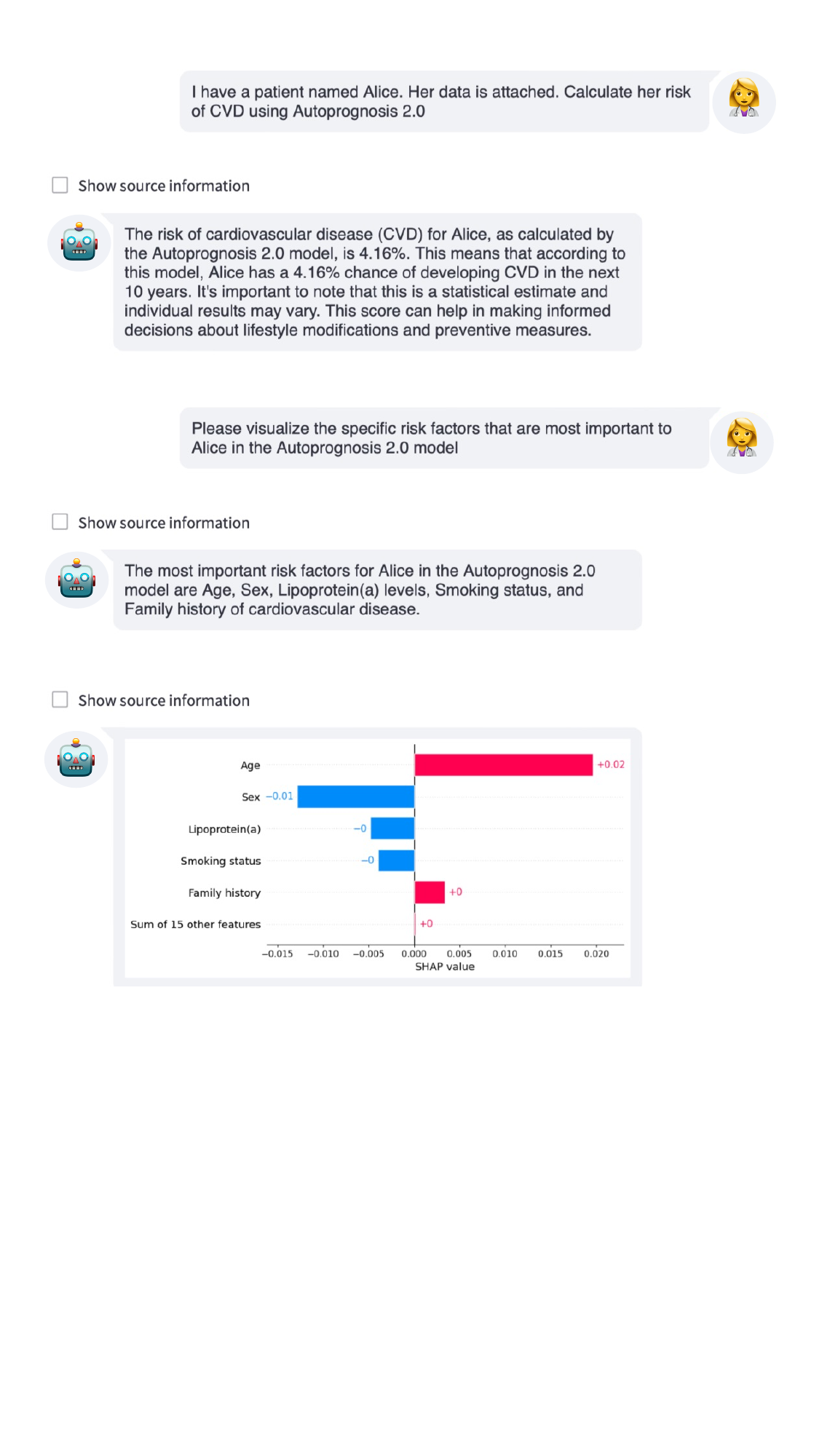}
  \caption{}
  \label{fig:eg_interaction_interpretability_1}
\end{subfigure}\hfill
\caption{\textbf{Example interactions with LLM-based interfaces.} (a) Clinicians can ask the LLM questions about the AutoPrognosis risk score, such as what variables are included, the underlying methodology, and the cohort on which the model was derived. (b) After calculating the patient's risk using the AutoPrognosis model, the clinician can query why this prediction was issued using explainable AI to improve understanding of the model predictions.}
\label{fig:eg_interactions_ap2}
\vspace{0.25cm}
\end{figure*}

\textbf{AutoPrognosis.}
We constructed an LLM-based interface for the machine learning model developed in this work (Fig. \ref{fig:eg_interactions_ap2}).
In our first example, we demonstrate the role such an interface could have upon model deployment (Fig. \ref{fig:eg_interaction_model_facts}. A clinician might have many questions about a risk score before using the predictions to inform recommendations or interventions (Table \ref{tbl:questions}). Here, the LLM provided specific information about the variables used by the risk score, the underlying methodology, how the model was constructed, and the cohort of individuals used to train the model.

Building model trust is a crucial step for prognostic models, in particular for models that are not inherently interpretable \cite{Rajpurkar2022}.
A recent study found that medical decision-makers had a strong preference for interactive explanations and, in particular, for these interactions to take the form of natural language dialogues \cite{lakkaraju2022rethinking}.
To enable this, in addition to allowing the LLM to access the model, we provided the LLM with the ability to apply methods from explainable AI (XAI) \cite{Imrie2023} to the underlying model, allowing users to investigate the rationale for predictions. 

\begin{figure}[!ht]
    \centering
    \includegraphics[width=0.59\linewidth,trim=10em 24.5em 10.85em 0.5em, clip]{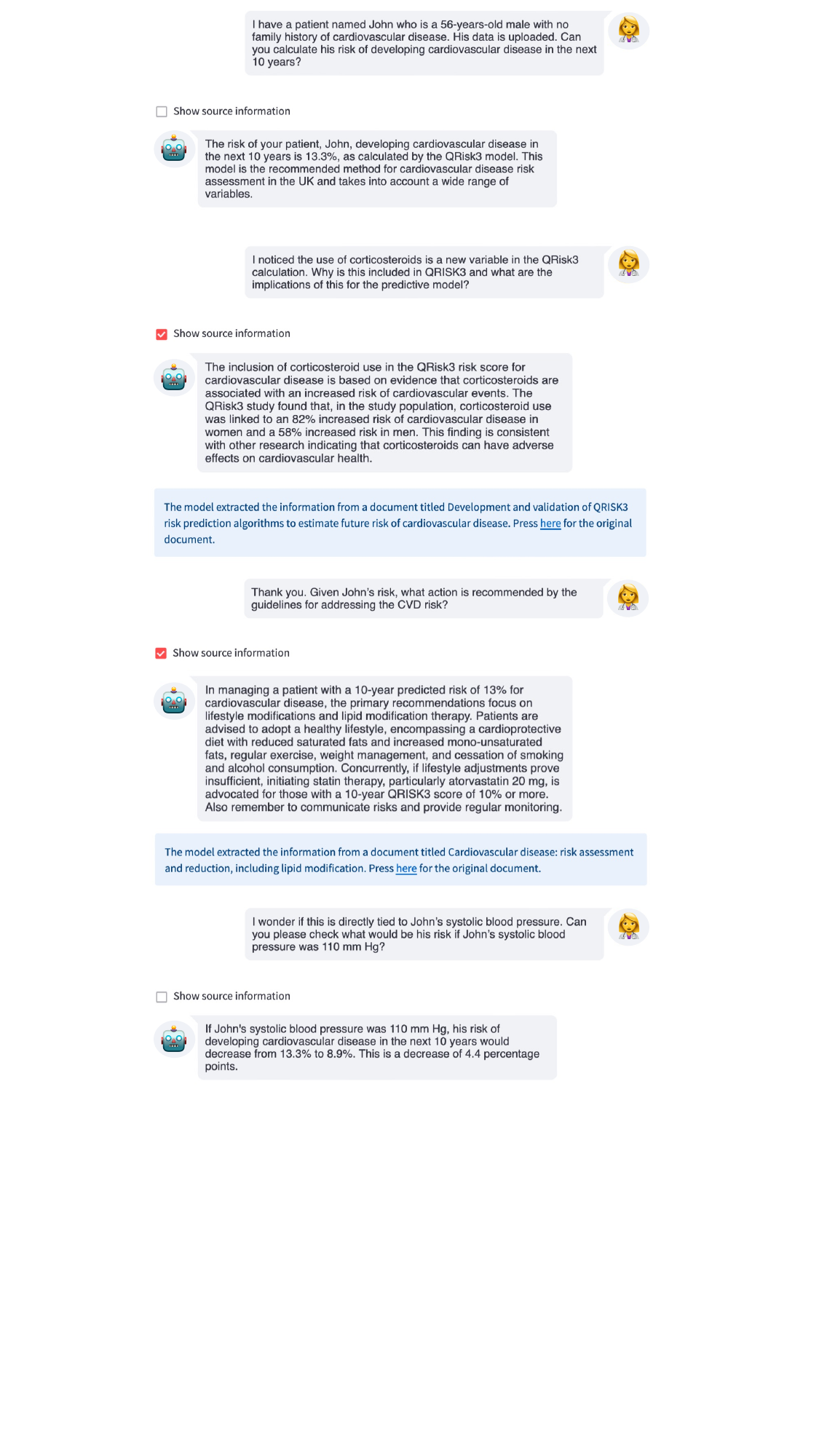}
    \caption{\textbf{Example interaction with an LLM-based interface for QRisk3.} The LLM uses QRisk3 to calculate the patient's 10-year risk of developing CVD, explains why certain features are included in the risk score using the QRisk3 paper \cite{Hippisley-Cox2017CVD}, and provides recommendations for this patient from the NICE clinical guidelines.}
    \label{fig:eg_interaction_qrisk}
\end{figure}

As shown in Fig. \ref{fig:eg_interaction_interpretability_1}, after calculating the patient's risk using the AutoPrognosis model, the clinician can query why this prediction was issued using XAI, facilitating a deeper understanding of the model. In this example, we used Shapley additive explanations (SHAP) \cite{lundberg2017unified} to explain the prediction. 
For this individual, their estimated 4.2\% risk was primarily caused by their age and family history of CVD, mitigated by being a woman, not smoking, and low levels of lipoprotein (a).
An additional interaction can be found in Fig. \ref{fig:eg_interaction_interpretability_2}. 
Due to their general pretraining, LLMs have knowledge of many topics; this can provide valuable additional information during interactions beyond the specific tools and external information sources provided to the LLM. 
For example, suppose the clinician is not familiar with the underlying XAI methodology, SHAP \cite{lundberg2017unified}. The LLM could explain how this approach works, in a variety of different ways and possibly over multiple interactions with the clinician, allowing specific queries or misunderstandings to be clarified.
This demonstrates the utility of LLMs beyond simply using existing tools and the benefit of their underlying knowledge.

\textbf{QRisk3.}
We also show how LLMs can incorporate existing tools and information for CVD risk prediction (Fig. \ref{fig:eg_interaction_qrisk}).
We provided the LLM access to QRisk3 \cite{Hippisley-Cox2017CVD}, a risk prediction tool that assesses the likelihood of developing CVD within 10 years.
Additionally, we provided the LLM with access to the academic paper describing QRisk3 \cite{Hippisley-Cox2017CVD} and the National Institute for Health and Care Excellence (NICE) clinical guidelines for CVD \cite{NICE_CVD}.
An example interaction between a physician and the LLM-based system is shown in Fig. \ref{fig:eg_interaction_qrisk}. An illustration of the reasoning process by which the LLM uses external tools is provided in Fig. \ref{fig:llm_reasoning_qrisk}.

In this example, when asked for the patient's 10-year risk of developing CVD, the LLM used QRisk3 to estimate the patient's risk, providing this to the user.
The LLM then summarized the QRisk3 paper to explain the inclusion of certain features before providing the recommended action for this patient from the NICE clinical guidelines. Finally, the LLM used QRisk3 to recalculate the patient's counterfactual risk assuming that they were able to reduce their systolic blood pressure to within normal ranges. This allows both the clinician and patient to understand the potential impact of changes to modifiable variables on the patient's risk.
While the clinician could have used the underlying resources to obtain this information, the LLM interface made the interaction simpler and more efficient, which has been identified as a key limitation of digital tools \cite{Ratwani2019decade,Mathews2019}.

\section{Discussion}

Large Language Models hold substantial promise for the medical domain, particularly in augmenting digital workflows and improving efficiency in healthcare delivery. 
The ability to integrate external tools and functionality with LLMs paves the way for innovative applications and can overcome limitations of LLMs, such as hallucinations. 
Doing so offers a potential transformation for how clinicians interact with digital tools and could help overcome the challenges of deploying clinical AI models.

We have demonstrated how LLMs can provide a unique interface between healthcare professionals and clinical predictive models, such as risk scores.
In particular, we developed a novel risk score for incidence of CVD using automated machine learning and developed LLM-based interfaces for our model and QRisk3 \cite{Hippisley-Cox2017CVD}, the current recommended risk score in the UK for CVD. 
Our approach is scalable and does not require any additional training of the language model, although approaches that improve with use could be yet more powerful.
Additionally, we aim to mitigate the problem of hallucination by ensuring that actionable advice is anchored in approved clinical resources, contrasting several previous applications of LLMs in medicine that focused exclusively on the knowledge and information learned by LLMs.

Numerous considerations exist before, during, and after assessing an individual's prognosis (Table \ref{tbl:questions}). 
Currently, clinicians must access these tools via fixed user interfaces or application programming interfaces (APIs), with existing interfaces typically only calculating risk.
Through an LLM-based interface, practitioners can obtain substantial additional information about the risk score, its development and methodology, the prediction issued, and related medical guidelines, in a manner that specifically addresses their needs or questions without providing superfluous information.

In this paper, we have focused on clinicians interacting with digital tools. However, there are numerous stakeholders in healthcare in addition to clinicians, such as patients, regulators, and administrators, each with different goals and requirements \cite{Imrie2023}. Our framework and approach could be applied to improve digital health interfaces for these alternate stakeholders. While this could have additional challenges, there are potentially even more substantial benefits for such individuals, given the differences in requirements, knowledge, and familiarity with digital health technology, among other factors.

While LLMs have general capabilities, they can lack domain-specific knowledge. This has led to the development of medical-focused LLMs, either by training new LLMs from scratch \cite{Luo2022,taylor2022galactica,Yang2022large} or by adapting existing general-purpose LLMs \cite{Singhal2023}. While we showed using such specialist LLMs is not required, they could be readily incorporated due to the modularity of our approach. This could further enhance the functionality of LLM-based interfaces.

As AI in medicine continues to advance, further research into LLMs and their potential applications in healthcare could provide significant benefits.
For example, LLMs could help to alleviate the data burden that is contributing to clinician burnout, as well as streamline patient management processes. 
Furthermore, studies have demonstrated high usability of LLMs, even with limited experience \cite{Skjuve2023user}, which is critical for successful clinical deployment.
While we believe this paper represents an important first step, we are only scratching the surface of the potential of LLMs in healthcare.
Ultimately, this line of work may significantly change the digital health landscape, enhancing the capabilities of clinicians and the quality of patient care.

\section{Methods}\label{sec:methods}

\subsection{Data source and study population}

We developed and validated our prognostic model using data from UK Biobank \cite{Sudlow2015}, a large prospective cohort of individuals from the UK. Participants in UK Biobank were enrolled between 2006 and 2010 and aged between 40 and 70 at the time of recruitment. 
We extracted a cohort of individuals with no known history of CVD at baseline. We excluded individuals who had been diagnosed with chronic kidney disease or type 1 diabetes and individuals who were being treated with statins.
This resulted in a cohort of 407,605 individuals (Fig. \ref{fig:flow_diagram}).

\subsection{Clinical Predictors and Outcomes}

We began by considering a number of predictors previously validated as CVD risk factors.
A subset of features were selected based on their predictive power and feature importance.
We included 20 features in our model, namely sex, age, atrial fibrillation, steroid treatment, type 2 diabetes, hypertension treatment, family history of CVD, systolic blood pressure, smoking status (never-smoker, ex-smoker, light smoker, moderate smoker, heavy smoker, alkaline phosphatase, apolipoprotein a, apolipoprotein b, cystatin c, C-reactive protein, HbA1c, IGF-1, lipoprotein a, triglycerides, urea, and overall health rating, which was self-reported by participants as excellent, good, fair or poor.

The primary outcome was incidence of CVD within a 10-year horizon. This was defined using the following ICD-10 codes: myocardial infarction (I21, I22), stroke (I63, I64), angina pectoris (I20), or transient cerebral ischaemic attacks (G45).
There were 17,600 CVD events within a 10-year horizon in the UK Biobank cohort.

\subsection{Model Derivation}
We trained a prognostic model using AutoPrognosis 2.0, an open-source automated machine learning software package \cite{Imrie2023AutoPrognosis}.
AutoPrognosis has been validated in several applications in medicine, for example to determine eligibility for lung cancer screening \cite{callender2023assessing}.
AutoPrognosis was used to optimize pipelines consisting of a variable preprocessing step followed by model selection and training. The optimized pipelines were subsequently combined in an ensemble using a weighted combination. 
Individual pipelines and the final ensemble were trained to maximize discrimination, measured using the area under the receiver operating curve. Predictions from the final ensemble were calibrated in a piecewise linear manner using the observed risk calculated using a Kaplan-Meier estimator \cite{kaplan1958nonparametric}. 

\subsection{Statistical Analysis}
Model discrimination was assessed using the concordance index (C-index) \cite{Uno2011} and calibration via calibration curves and the ratio between the expected and observed number of events. We also assessed Brier scores \cite{brier1950verification}.
Calibration curves were constructed by first dividing the population into 10 risk deciles based on each individual's predicted 10-year risk and then comparing the mean estimated risk to the observed risk in each decile. 
Kaplan–Meier estimators \cite{kaplan1958nonparametric} were used to calculate observed risk.
Additionally, we conducted decision curve analysis, assessing the net benefit for a range of risk thresholds.
In addition to the existing risk scores, when assessing net benefit, we also compared our model with baseline strategies that assume all patients develop CVD (All) or none do (None).
Results are based on 5-fold cross-validation.
We used multiple imputation with chained equations (MICE) \cite{vanBuuren2011} to impute missing values. We generated five imputed datasets and combined results using Rubin's rules \cite{rubin2018multiple}.

\subsection{Model Comparisons}
We compared our model with three existing risk scores, namely Framingham score \cite{DAgostino2008}, SCORE2 \cite{SCORE2}, and QRisk3 \cite{Hippisley-Cox2017CVD}. All three models assess the likelihood of developing CVD within a 10 year horizon.

The Framingham score \cite{DAgostino2008} was recommended by the 2010 American College of Cardiology/American Heart Association (ACC/AHA) guidelines \cite{Greenland2010} and is based on eight risk factors: sex, age, systolic blood pressure, treatment for hypertension, smoking status, history of diabetes, total cholesterol, and high-density lipoprotein (HDL) cholesterol.

SCORE2 \cite{SCORE2} was developed using data from 45 cohorts in 13 countries across Europe, including the United Kingdom, and has been included in the European Society of Cardiology guidelines. 
SCORE2 employs seven risk factors: sex, age, systolic blood pressure, smoking status, history of diabetes, total cholesterol, and high-density lipoprotein (HDL) cholesterol.

QRisk3 \cite{Hippisley-Cox2017CVD} was developed on data from 981 practices in England comprising almost 8 million patients and is currently the recommended risk score for CVD in England  \cite{NICE_CVD}.
QRisk3 uses 22 variables, including demographics (such as sex, age), lifestyle factors (e.g. smoking), medical and family history (e.g. history of diabetes, family history of CVD), medical examination (e.g. blood pressure), and blood tests (total cholesterol/HDL cholesterol ratio).

\subsection{LLM-based interface}

Here, we describe the implementation details for the LLM-based systems demonstrated above.

\subsubsection{LLM framework}

By default, LLMs will provide responses in the form of text based on the provided context, such as a prompt or conversation history.
To construct interfaces for digital tools using LLMs, we instead viewed the LLM as an agent that can interact with an environment to solve tasks. 
Formally, at each step $t \in T$, the agent receives observation $o_t \in O$ from the environment and subsequently takes action $a_t \in A$ according to policy $\pi(a_t \vert h_t)$, where $h_t = (o_0, a_0, \dots , o_{t-1}, a_{t-1}, o_t)$ is the history.
To enable the agent to both reason and use external tools, we used the ReAct framework \cite{yao2022react} which decomposes the action space as $\hat{A} = A \union L$, where $a \in A$ are actions using specific tools and an action $a \in L$ represents not using an external tool but instead allows the model to reason over the history about what action to take next. 

Since we will provide the agents tasks in the form of natural language, and actions in the language space $L$ are (essentially) infinite, we chose to benefit from strong language priors and use a pretrained LLM.
We implemented our LLM-based interfaces using GPT-4 \cite{openai2023gpt4}, with access through the OpenAI API. Interactions with external tools were implemented using LangChain \cite{LangChain}. 

Frameworks such as Toolformer \cite{schick2023toolformer} and WebGPT \cite{nakano2021webgpt} trained LLMs to use basic tools, such as calculators, calendars, and search engines, via self-supervised fine-tuning and fine-tuning using behavior cloning and reinforcement learning, respectively. 
In contrast, following ReAct \cite{yao2022react}, we used in-context learning \cite{dong2023survey} in the form of prompting, providing the LLM with sufficient information about possible actions and using the underlying reasoning capabilities of the LLM.
This removes the need for further training of the LLM, which might be challenging in the medical domain without suitable examples. This also enabled us to use multiple tools.
We built user interfaces using StreamLit \cite{streamlit}.

\subsubsection{External Tools}

This section describes the external tools and sources of information made available to the LLM in our examples.

\textbf{LLM-based interface for AutoPrognosis.}
To demonstrate an LLM-based interface, we equipped the LLM with the CVD risk prediction model developed in this work on the UK Biobank cohort using AutoPrognosis 2.0 \cite{Imrie2023AutoPrognosis}.
Building model trust is a crucial step for prognostic models, in particular for models that are not inherently interpretable \cite{Rajpurkar2022,asan2020artificial}.
A recent study found that medical decision-makers had a strong preference for interactive explanations and, in particular, for these interactions to take the form of natural language dialogues \cite{lakkaraju2022rethinking}.
To enable this, we enabled the LLM to use explainable AI (XAI) methods \cite{Imrie2023} on the underlying model, allowing users to investigate the rationale for predictions, both in general and for the specific patient. 
In particular, we used SHAP \cite{lundberg2017unified} to interpret model predictions.
We additionally provided the LLM with information about the variables used by the risk score, the underlying methodology and how the model was constructed, and details regarding the cohort used to develop the model.
Example interactions are shown in Fig. \ref{fig:eg_interactions_ap2} and Fig. \ref{fig:eg_interaction_interpretability_2}.

\textbf{LLM-based interface for QRisk3.}
As a second example, we showed how LLMs can incorporate existing tools and information for CVD risk prediction.
We provided the LLM access to QRisk3 \cite{Hippisley-Cox2017CVD} and enabled the LLM to use the risk score either using the provided data or, if requested by the user, to modify a variable and assess the impact of such a change on the patient's risk.
Additionally, we provided the LLM with access to the academic paper describing QRisk3 \cite{Hippisley-Cox2017CVD} and the National Institute for Health and Care Excellence (NICE) clinical guidelines for CVD \cite{NICE_CVD}. 
An example interaction is shown in Fig. \ref{fig:eg_interaction_qrisk}.

\section{Code availability}\label{sec:code}

Code for the LLM-based interfaces can be accessed at \url{https://github.com/pauliusrauba/LLMs_interface}.
AutoPrognosis is an open-source package available on GitHub at \url{https://github.com/vanderschaarlab/AutoPrognosis} and on PyPI at \url{https://pypi.org/project/autoprognosis/}.

\section{Data availability}\label{sec:data}

This research has been conducted using the UK Biobank resource under application number 105160. 
Data from UK Biobank is accessible through a request process (\url{https://www.ukbiobank.ac.uk/enable-your-research/register}). The authors had no special access or privileges when accessing the data.

\backmatter

\bmhead{Acknowledgments}
This study received no funding.

\bmhead{Competing interests}
All authors declare no financial or non-financial competing interests. 

\bmhead{Author contributions}
F.I. and M.vdS. conceptualized the manuscript. F.I. and P.R. designed and performed the experiments. F.I. wrote the original draft of the manuscript, and all authors contributed to editing and revising it. 

\bibliography{bibliography}

\clearpage

\begin{appendices}

\renewcommand{\thefigure}{S.\arabic{figure}}
\setcounter{figure}{0}

\renewcommand{\thetable}{S.\arabic{table}}
\setcounter{table}{0}

\begin{figure*}[ht!]
  \centering
  \includegraphics[width=0.6\linewidth,trim=2em 5.5em 11em 5em, clip]{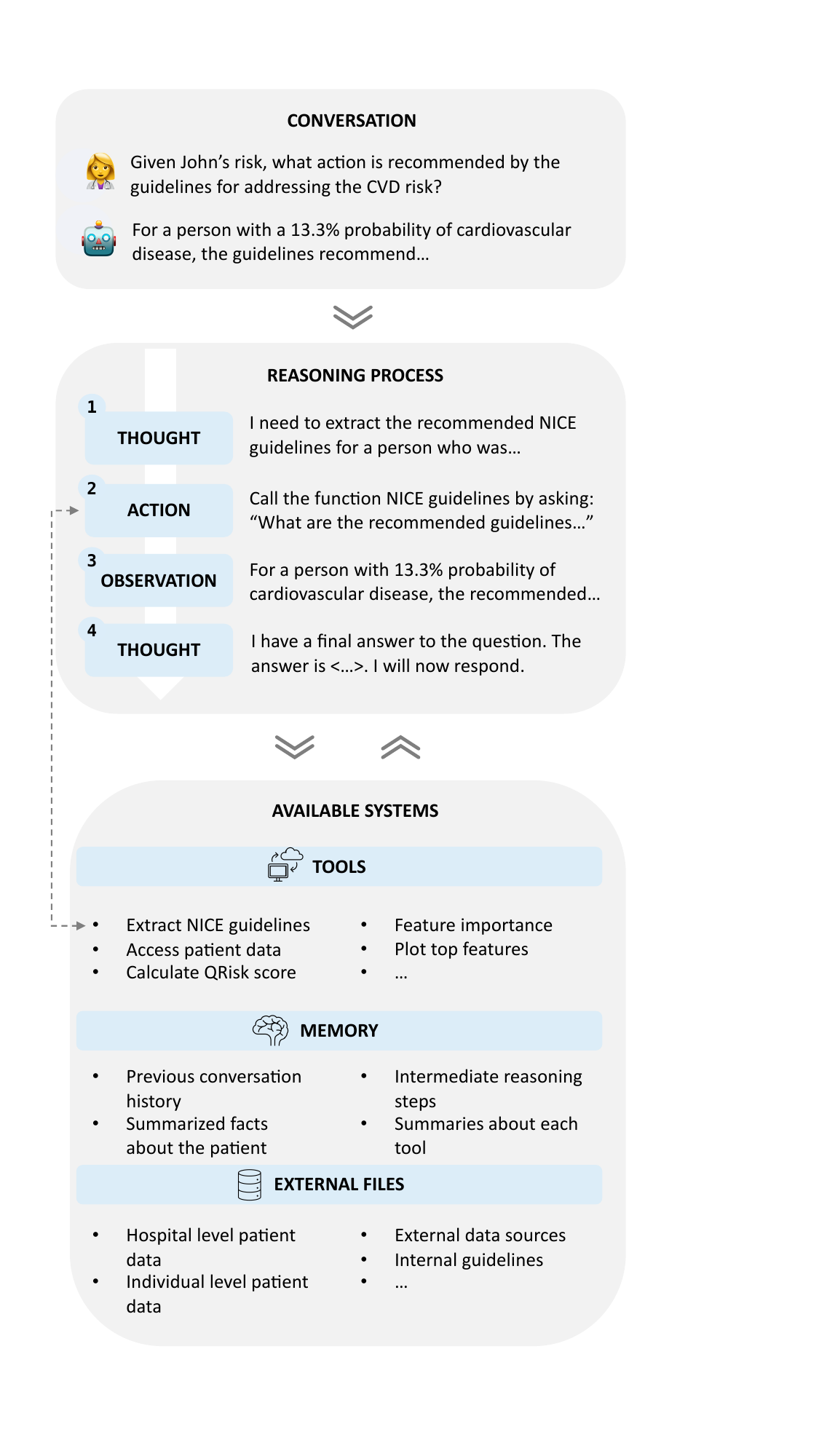}
  \caption{Illustration of the process by which the LLM uses external tools. The LLM is provided with the history of the interaction, including the current request. Using an iterative reasoning process, the LLM decides which, if any, tools are required and with what input (``Thought''). The LLM then uses the external tool (``Action'') and receives the output (``Observation''). Finally, the LLM decides to answer the question (``Thought''), providing a response to the user.}
  \label{fig:llm_reasoning_qrisk}
\end{figure*}

\begin{figure}[!th]
    \centering
    \includegraphics[width=0.8\linewidth]{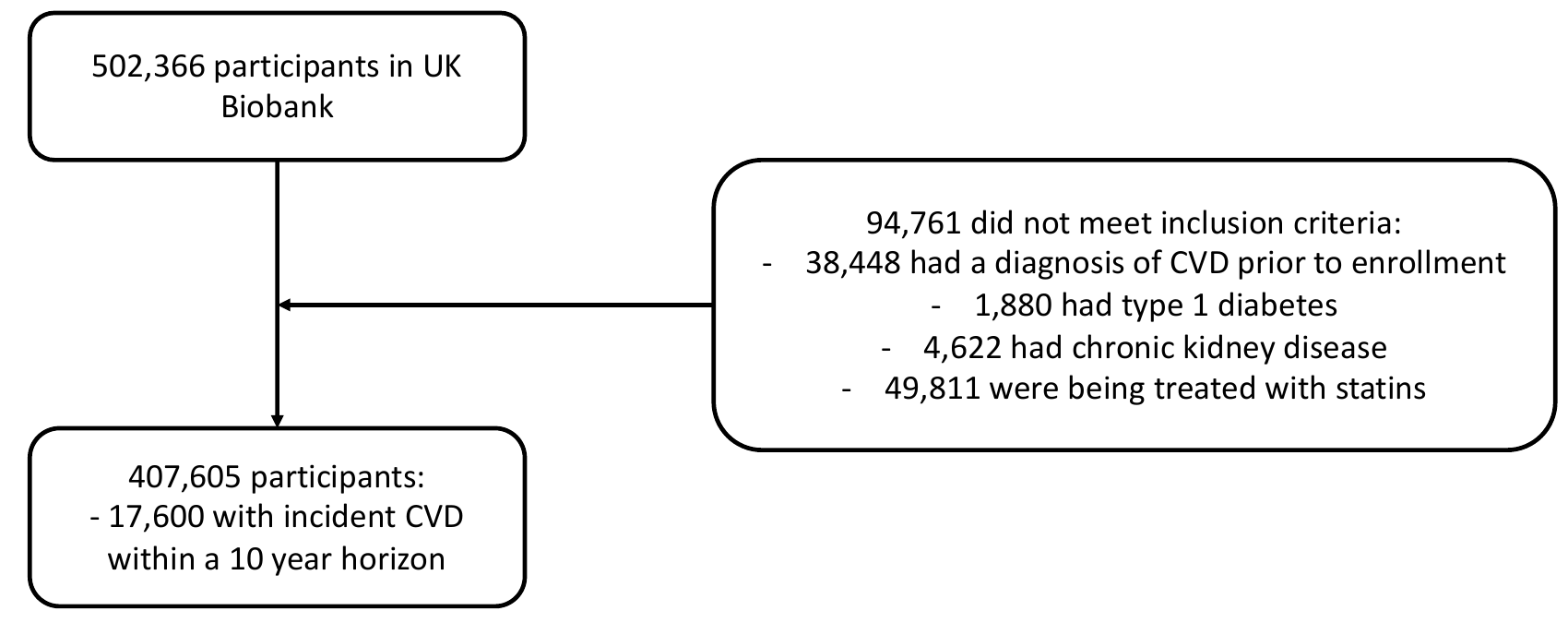}
    \caption{Flow diagram of UK Biobank participants.}
    \label{fig:flow_diagram}
\end{figure}

\begin{table}[ht!]
\caption{Descriptive characteristics of the UK Biobank cohort.}
\label{tbl:cohort_characteristics}
\vspace{0.1in}
\centering
\begin{tabular}{lccc}
\toprule
                        & \textbf{All}         & \textbf{CVD events}         & \textbf{No CVD event}         \\
                        & \textbf{(n=407,605)} & \textbf{(n=17,600, 4.32\%)} & \textbf{(n=390,005, 95.68\%)} \\
\hline
\textbf{Age}            &                   &                      &                        \\
\textless{}50           & 110,750 (27.17\%) & 1,876 (10.66\%)      & 108,874 (27.92\%)      \\
50-60                   & 142,784 (35.03\%) & 5,272 (29.95\%)      & 137,512 (35.26\%)      \\
$\geq$60                & 154,071 (37.80\%) & 10,452 (59.39\%)     & 143,619 (36.82\%)      \\
\textit{Missing}        & 0 (0.00\%)        & 0 (0.00\%)           & 0 (0.00\%)             \\
                        &                   &                      &                        \\
\textbf{BMI}            &                   &                      &                        \\
\textless{}18.5         & 2,403 (0.59\%)    & 85 (0.48\%)          & 2,318 (0.59\%)         \\
18.5-25                 & 144,179 (35.37\%) & 4,514 (25.65\%)      & 139,665 (35.81\%)      \\
25-30                   & 171,140 (41.99\%) & 7,978 (45.33\%)      & 163,162 (41.84\%)      \\
30-35                   & 87,529 (21.47\%)  & 4,880 (27.73\%)      & 82,649 (21.19\%)       \\
\textit{Missing}        & 2,354 (0.58\%)    & 143 (0.81\%)         & 2,211 (0.57\%)         \\
                        &                   &                      &                        \\
\multicolumn{2}{l}{\textbf{Systolic Blood Pressure}} &             &                        \\
\textless{}120          & 89,725 (22.01\%)  & 2,027 (11.52\%)      & 87,698 (22.49\%)       \\
120-140                 & 164,263 (40.30\%) & 6,157 (34.98\%)      & 158,106 (40.54\%)      \\
\textgreater{}140       & 128,984 (31.64\%) & 8,209 (46.64\%)      & 120,775 (30.97\%)      \\
\textit{Missing}        & 24,633 (6.04\%)   & 1,207 (6.86\%)       & 23,426 (6.01\%)        \\
                        &                 &                        &                        \\
\multicolumn{2}{l}{\textbf{Total Cholesterol / HDL}} &             &                        \\
\textless{}4.5          & 224,495 (55.08\%) & 7,282 (41.38\%)      & 217,213 (55.69\%)      \\
4.5-6                   & 98,739 (24.22\%)  & 5,747 (32.65\%)      & 92,992 (23.84\%)       \\
\textgreater{}6         & 25,215 (6.19\%)   & 2,075 (11.79\%)      & 23,140 (5.93\%)        \\
\textit{Missing}        & 59,156 (14.51\%)  & 2,496 (14.18\%)      & 56,660 (14.53\%)       \\
                        &                   &                      &                        \\
\textbf{Sex}            &                   &                      &                        \\
Men                     & 173,818 (42.64\%) & 10,360 (58.86\%)     & 163,458 (41.91\%)      \\
Women                   & 233,787 (57.36\%) & 7,240 (41.14\%)      & 226,547 (58.09\%)      \\
\textit{Missing}        & 0 (0.00\%)        & 0 (0.00\%)           & 0 (0.00\%)             \\
                        &                   &                      &                        \\
\textbf{Smoking}        &                   &                      &                        \\
Never                   & 238,095 (58.41\%) & 8,608 (48.91\%)      & 229,487 (58.84\%)      \\
Previous                & 125,829 (30.87\%) & 6,202 (35.24\%)      & 119,627 (30.67\%)      \\
Current                 & 41,420 (10.16\%)  & 2,646 (15.03\%)      & 38,774 (9.94\%)        \\
\textit{Missing}        & 2,261 (0.55\%)    & 144 (0.82\%)         & 2,117 (0.54\%)         \\
\bottomrule
\end{tabular}
\end{table}

\begin{figure}[!ht]
    \centering
    \includegraphics[width=0.6\linewidth,trim=1.75em 31em 2.75em 1em, clip]{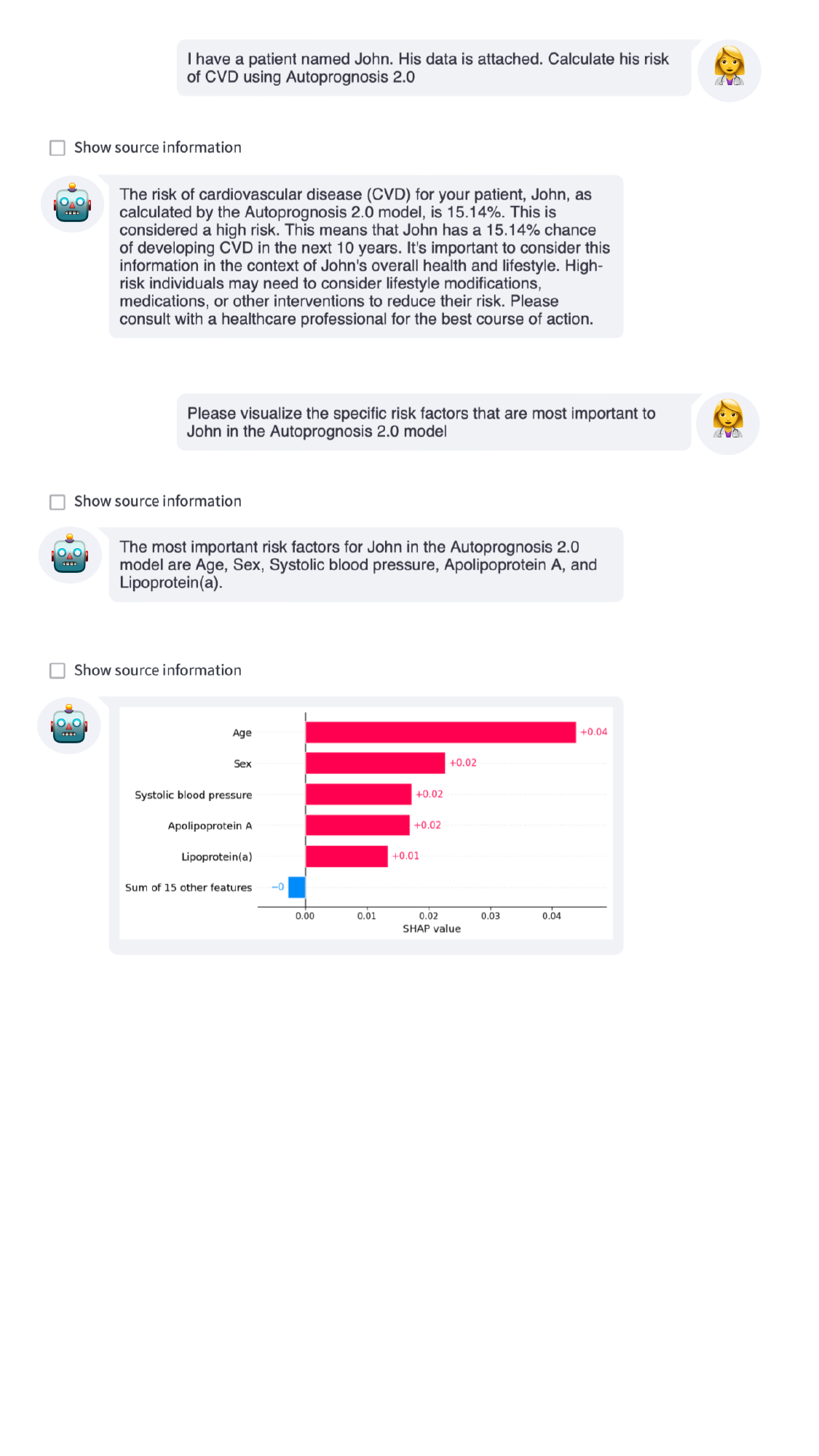}
    \caption{\textbf{Example interaction with an LLM-based interface.} After calculating the patient's risk using the AutoPrognosis model, the clinician can query why this prediction was issued using explainable AI.}
    \label{fig:eg_interaction_interpretability_2}
\end{figure}

\end{appendices}

\end{document}